%% file: ICCV_C3.tex
\documentclass[10pt,twocolumn,letterpaper]{article}

\usepackage{iccv}
\usepackage{times}
\usepackage{epsfig}
\usepackage{graphicx}
\usepackage{amsmath}
\usepackage{amssymb}

\usepackage[dvipsnames]{xcolor}
\usepackage{tabularx}
\usepackage{multicol}
\usepackage{array}

\newif\ifdraft\draftfalse
\ifdraft

\definecolor{greyblue}{rgb}{0.1,0.6,0.5}
\newcommand\dy[1]{\textcolor{greyblue}{#1}}
\newcommand\yj[1]{\textcolor{blue}{#1}}

\newcommand\hj[1]{\textcolor{Sepia}{#1}}
\newcommand\nj[1]{\textcolor{magenta}{#1}}
\newcommand\gs[1]{\textcolor{RoyalBlue}{#1}}
\else

\newcommand\dy[1]{#1}
\newcommand\yj[1]{#1}

\newcommand\hj[1]{#1}
\newcommand\nj[1]{#1}
\newcommand\gs[1]{#1}
\fi
  {\begin{list}{}%
          {\setlength{\leftmargin}{#1}}%
          \item[]%
  }
  {\end{list}}

\usepackage[pagebackref=true,breaklinks=true,letterpaper=true,colorlinks,bookmarks=false]{hyperref}

\iccvfinalcopy 

\def\iccvPaperID{34} 

\ificcvfinal\fi

\begin{document}

\title{C3: Concentrated-Comprehensive Convolution and \\
its application to semantic segmentation}

\author{
Hyojin Park \textsuperscript{1}\\
{\tt\small wolfrun@snu.ac.kr}
\and
Youngjoon Yoo \textsuperscript{2}\\
{\tt\small youngjoon.yoo@navercorp.com}
\and
Geonseok Seo \textsuperscript{1}\\
{\tt\small geonseoks@snu.ac.kr}
\and
Dongyoon Han \textsuperscript{2}\\
{\tt\small dongyoon.han@navercorp.com}
\and
Sangdoo Yun \textsuperscript{2}\\
{\tt\small sangdoo.yun@navercorp.com}
\and
Nojun Kwak \textsuperscript{1}\\
{\tt\small nojunk@snu.ac.kr}
\and
\textsuperscript{1} Seoul National University 
\textsuperscript{2} CLOVA AI Research, Naver Corp.
}

\maketitle
\ificcvfinal\thispagestyle{empty}\fi

\begin{abstract}
  One of the practical choices for making a lightweight semantic segmentation model is to combine a depth-wise separable convolution with a dilated convolution.
However, the simple combination of these two methods results in an over-simplified operation which causes severe performance degradation due to loss of information contained in the feature map.
To resolve this problem, we propose a new block called Concentrated-Comprehensive Convolution (C3) which applies the asymmetric convolutions before the depth-wise separable dilated convolution to compensate for the information loss due to dilated convolution.
The C3 block consists of a concentration stage and a comprehensive convolution stage.
The first stage uses two depth-wise asymmetric convolutions for compressed information from the neighboring pixels to alleviate the information loss. 
The second stage increases the receptive field by using a depth-wise separable dilated convolution from the feature map of the first stage. 
We applied the C3 block to various segmentation frameworks (ESPNet, DRN, ERFNet, ENet) for proving the beneficial properties of our proposed method.
Experimental results show that the proposed method preserves the original accuracies on Cityscapes dataset while reducing the complexity.
Furthermore, we modified ESPNet to achieve about 2\% better performance while reducing the number of parameters by half and the number of FLOPs by 35\% compared with the original ESPNet.
\nj{Finally, experiments on ImageNet classification task show that C3 block can successfully replace dilated convolutions.}
\end{abstract}


\input{Eng/Intro.tex}

\input{Eng/Related.tex}
\input{Eng/Method.tex}
\input{Eng/Experiment.tex}
\input{Eng/Conclusion.tex}


\begin{table*}[t]
\centering
    \begin{tabular*}{0.99\textwidth}{@{\extracolsep{\fill}} ccc}
    \centering{Layer} & Operation & Flop \\
     \hline \hline
    {Convolution}& $O=F*K$  & $2\cdot H_oW_o\cdot K_hK_w\cdot C_iC_o/g$ \\
     \hline
  {Deconvolution}& $O=F*K$  &  $2\cdot H_iW_i\cdot K_hK_w\cdot C_iC_o/g$ \\
     \hline
     {Average Pooling}& $O=Avg(F)$ & $H_i\cdot W_i\cdot C_i$ \\
     \hline
    Bilinear upsampling & $f(x,y)=\sum_{i=0}^{1}\sum_{j=0}^{1}a_{ij}x^iy^j$ 
    & $3\cdot H_i\cdot W_i\cdot C_i$  \\
     \hline
      {Batch normalization}& $(F-mean)/std$ &  $ 2\cdot H_i\cdot W_i\cdot C_i$ \\
     \hline
    {ReLU or PReLU}& $O=g(F)$  & $ H_i\cdot W_i\cdot C_i$ \\
     \hline
 
    \end{tabular*}%

      \caption{The detail method for calculating FLOPs}
  \label{tab:flop}%
\end{table*}%

{\small
\bibliographystyle{ieee}
\bibliography{egbib}
}
\newpage

\section*{Appendix}
\subsection*{A. Flops Calculation}
\label{flop}

\nj{Table \ref{tab:flop} shows how we calculated FLOPs for each operation.}
The following notations are used.  \newline
$F$ : A input feature map \newline
$O$ : A output feature map \newline
$K$ : A convolution kernel \newline
$K_h$ : A height of convolution kernel \newline
$K_w$ : A width of convolution kernel \newline
$H_i$ : A height of input feature map \newline  
$W_i$ : A width of input feature map \newline 
$C_i$ : A input channel dimension of feature map or kernel \newline
$C_o$ : A output channel dimension of feature map or kernel \newline
$g$ : A group size for channel dimension \newline
$H_o$ : A height of output feature map \newline
$W_o$ : A width of output feature map \newline
$g(\cdot)$ : A non-linear activation function\newline

\end{document}

%% file: Eng/Intro.tex

\section{Introduction}
\label{sec:intro}

  \begin{figure}[h]
\small
    \centering
  \begin{tabular}{c}
    \includegraphics[width=0.97\linewidth]{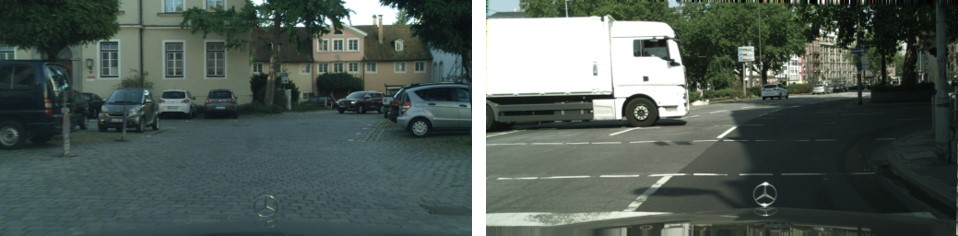} \\
    (a) Input image\\
     \includegraphics[width=0.97\linewidth]{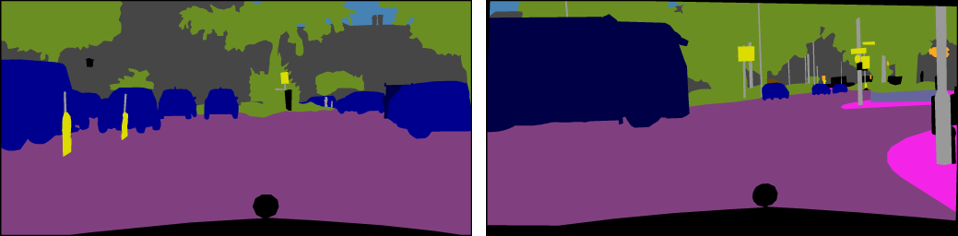}  \\
    (b) Ground Truth\\
     \includegraphics[width=0.97\linewidth]{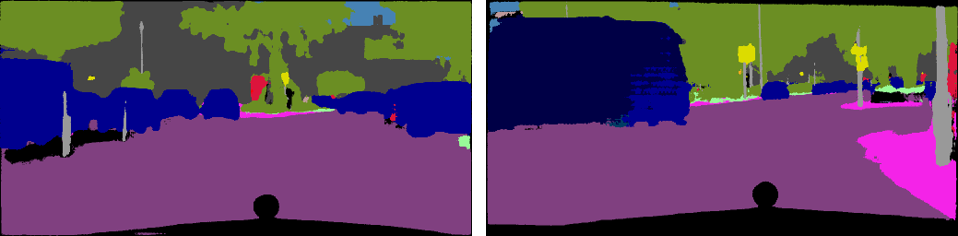}  \\
    (c) Result of ESPNet (Param : 0.364M)\\
     \includegraphics[width=0.97\linewidth]{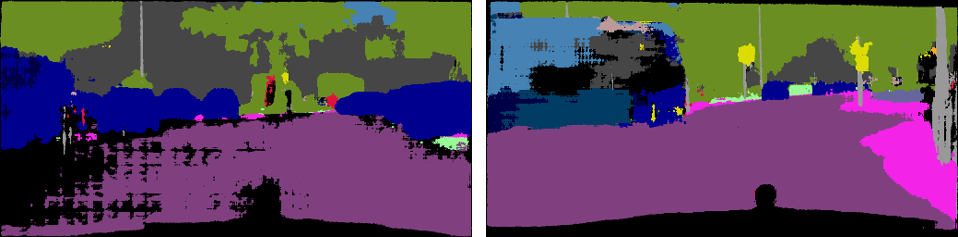}  \\
    (d) Result of ds-Dilate (Param : 0.187M)\\
     \includegraphics[width=0.97\linewidth]{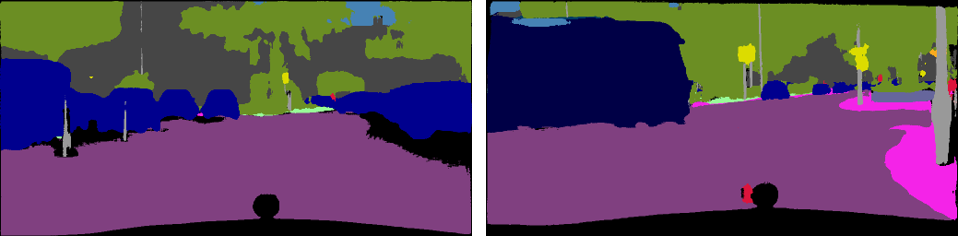}  \\
    (e) Result of our C3 block (Param : 0.198M)\\
    \end{tabular}%

    \caption{ Illustrations of performance degradation from depth-wise separable dilated convolution \textit{(ds-Dilate)}. 
    \nj{(c): Original ESPNet  \cite{mehta2018espnet}, (d): ESPNet with increased layers using ds-Dilate, (e): Our model using C3 blocks with the same encoder-decoder structure as \hj{(c)}. Param denotes the number of parameters for each model.}
    \nj{The numbers are} the number of parameters and mIOU calculated by the test set on Cityscapes benchmark.
    }
    \label{fig:teaser}
\end{figure}

Deep network-based semantic segmentation algorithms \nj{have significantly} enhanced the performance, but \nj{they demand} heavy \nj{computational costs}.
It is basically because semantic segmentation is a pixel-wise classification \nj{problem}.
Most of the \nj{semantic segmentation models usually have} an encoder-decoder structure \nj{where the} 
encoder reduces \nj{the size of the} feature map to \nj{obtain a large enough} receptive field, \nj{while the} decoder recovers original resolution and spatial information from the small\nj{-}size\nj{d} feature map.
Recent segmentation studies have actively used dilated convolutions \nj{to obtain} a wide receptive field without increasing the number of parameters and \nj{to} reduce the burden of the decoder 
while enhancing \nj{the} speed and accuracy.
However, even though many models utilized dilated convolutions, \nj{some models are still heavy} for an embedded system.
\nj{There have been other} kinds of researches \nj{which} have made efforts to develop \nj{a} lightweight \nj{model}~\cite{romera2018erfnet, paszke2016enet, poudel2018contextnet, siam2018rtseg}. 
Among \nj{the lightweight models}, 
ESPNet~\cite{mehta2018espnet} 
\yj{shows the \nj{best} segmentation performance} 
by adopting a well-designed series of spatial pyramid dilated convolutions.

For further slimming the segmentation network \nj{towards} embedded system\nj{s},
a practical choice \nj{would be to apply the} depth-wise separable convolution \textit{(ds-Conv)} \cite{chollet2017xception, howard2017mobilenets}, which is \nj{a} popular method to reduce the number of parameters and the \nj{computational complexity}, \nj{to the dilated convolution}.
However, we observed that this combination commonly leads to huge performance degradation as shown in Figure \ref{fig:teaser}, and there is no in-depth analysis of this problem \nj{so far}.
Figure \ref{fig:teaser} (c), (d) and (e) use different basic convolution blocks in \nj{an} ESPNet-based structure.
Figure \ref{fig:teaser} (c) and (e) are the results of ESPNet-based networks with the same encoder-decoder structure, but (d) increases the number of layers in the encoder for similar complexity with (e). 
Figure \ref{fig:teaser} (d) is obtained using simple depth-wise separable dilated convolution \textit{(ds-Dilate)} blocks, which has less accuracy than ours.
\nj{The reason of this performance degradation can be attributed to 1) inexact approximation of the standard convolution with the ds-Conv and 2) information loss on the neighboring pixels in a feature map due to the use of the dilated convolution.} The combination of \nj{the} two method\nj{s further} leads to \nj{a} too\nj{-}sparse operation and loss of information.
Our proposed method resolve\nj{s} this problem and \nj{Figure \ref{fig:teaser}(e) shows an improved} segmentation quality \nj{compared with} Figure \ref{fig:teaser}(d). 

In this paper, we \nj{introduce} \yj{a} new concentrated-comprehensive convolution (C3\footnote{We changed the method name from CCC to C3.}) block for reducing the complexity of segmentation model\nj{s} while preserving the original accuracy.
\nj{By replacing} a dilated convolution \nj{with} our proposed method, \nj{we} resolve the problem of naive ds-Dilate with \nj{reduced computational complexity}.
Our main contributions can be summarized as follows: 
(1) We propose a new convolutional block \nj{which can replace the standard dilated convolution to reduce the complexity of a segmentation model} while preserving accuracy.
Our proposed block can \nj{be} applied to \nj{form a} lightweight model. 
(2) We conducted extensive experiments \nj{on Cityscapes} to prove \nj{that} our proposed C3 block can be easily applied \yj{in} a plug-and-play \yj{manner} to \yj{many} models (ESPNet, ERFNet, ENet and DRN) which use dilated convolutions. 
Furthermore, \nj{the proposed block is applied not only to the segmentation task but also to the classification task
\yj{by adapting the proposed block} to DRN \cite{Yu2017Drn}.} 
(3) We propose a better lightweight model based on ESPNet \yj{can be developed} by using \yj{the} proposed C3 block. 
The re-designed model 
reduces the number of parameters by almost half, and the computational cost in FLOPs by 35\% compared to the original ESPNet.
\yj{Also, this model} achieve\yj{s} better accuracy than ESPNet \nj{with} real-time execution (16.4 FPS) in an embedded board (Nvidia-TX2 board).
\hj{(4) Qualitatively, we showed \nj{that} ds-Dilate \nj{was unable to} make enough feature activation due to loss  of information \nj{and  demonstrated} that our proposed C3 block could recover this feature activation like the original model \nj{using the} heatmap from Grad-CAM \cite{selvaraju2017grad}.
}



%% file: Eng/Related.tex
\section{Related Work}
\label{sec:related}

\noindent
\textbf{Convolution Factorization: } 
Convolution factorization divides a convolution operation into several stages to reduce the complexity than original convolutions.
In Inception \cite{szegedy2015going, szegedy2016rethinking, szegedy2017inception}, several convolutions were performed respectively, and the results were concatenated. Then, a $1\times 1$ convolution was used to reduce the number of channels. 
Xception \cite{chollet2017xception}, MobileNet \cite{howard2017mobilenets} and  MobileNetV2 \cite{sandler2018inverted} used the depth-wise separable convolution \textit{(ds-Conv)}, which performs spatial and cross-channel operations separately for decreasing computation.
ResNeXt \cite{xie2017aggregated} and ShuffleNet \cite{zhang2017shufflenet} applied a group convolution to reduce complexity.

\noindent
\textbf{Dilated Convolution: }
Dilated convolution is a widely used method in segmentation due to a large receptive field without increasing the amount of complexity by inserting holes between the consecutive kernel pixels. 
DeepLab V3 \cite{chen2017rethinking} and DenseASPP \cite{yang2018DenseAspp} used a set of different dilation rate convolutional layers to generate more dense multi-scale feature representation.
DRN \cite{Yu2017Drn} added dilated convolution in ResNet \cite{he2016deep} and modified the residual connection for better accuracy.
Recent researches have started to combine dilated convolution with ds-Conv.
In machine translation, \cite{kaiser2018depthwise} proposed a super-separable convolution that divides the channel dimension of a tensor into several groups and performs ds-Conv for each.
However, they did not solve performance degradation from combining the dilated convolution and the ds-Conv.
DeepLabv3+ \cite{chen2018encoder} applies the ds-Conv to the atrous spatial pyramid pooling (ASPP) and a decoder module based on Resnet \cite{he2016deep} and Xeception model \cite{chollet2017xception}.

\noindent
\textbf{Lightweight Segmentation: }
Enet \cite{paszke2016enet} is the first architecture designed for real-time segmentation. Since then, ESPNet \cite{mehta2018espnet} has led to more improvements than ever before in both the speed and the performance by an efficient spatial pyramid of dilated convolutions.
ERFNet \cite{romera2018erfnet} used residual connections and factorized a dilated convolution.
The authors of \cite{vallurupalli2018efficient} proposed a new training method which gradually changes a dense convolution to a grouped convolution and led to a $5$-times improvement in speed than ERFNet  \cite{romera2018erfnet}
ContextNet \cite{poudel2018contextnet} designed two network branches for global context and detail information.

Combining dilated convolution and ds-Conv can reduce computation, but the simple combination of these induces severe degradation in accuracy, and there is no careful consideration of this problem. 
Our method resolves this problem from the concept of minimizing information loss by replacing the dilated convolutions with the newly proposed concentrated-comprehensive convolutions (C3) blocks. 
The proposed C3 block is not restricted to a specific model, but it can be applied to other models which use dilated convolutions.
\vspace{-2mm}

%% file: Eng/Method.tex


\begin{figure}[t]
    \centering
    \begin{tabular}{c}
          \includegraphics[width=0.97\linewidth]{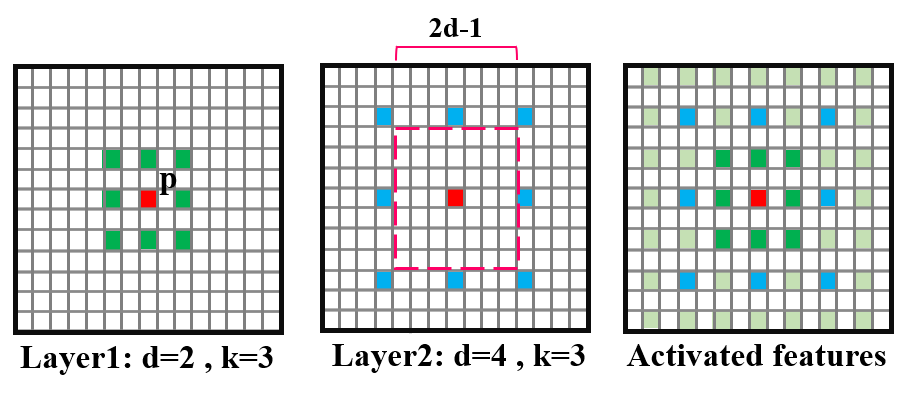}  \\
         (a) Local information missing  \\
             \includegraphics[width=0.97\linewidth]{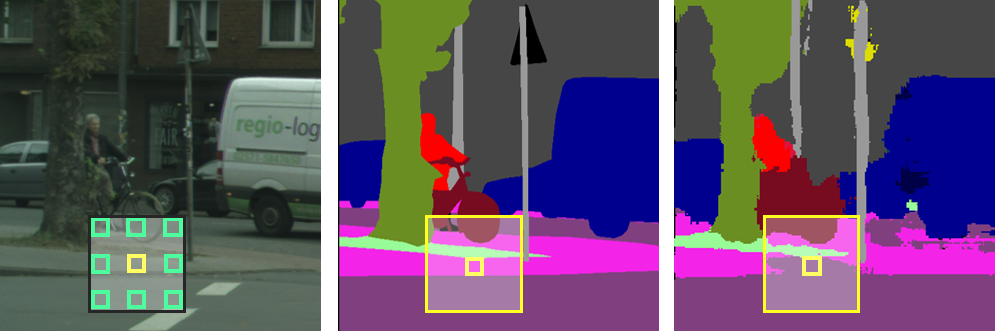}  \\
         (b) Impact of sparse computation  \\
    \end{tabular}
    \caption{\textbf{ \hj{Two negative cases from dilated convolution}}. (a) In two consecutive layers of dilated convolution, features in white are not used for computing the output feature in the center, marked in red. (b) The pixels in the green boxes influence the segmentation result of the area in the yellow box. From left to right, an input image, a ground truth and an estimated label image}
    \label{fig:dilation}
\end{figure}

\section{Method}
\label{sec:Model}

\begin{figure*}[t]
  \centering
 \includegraphics[{width=0.97\linewidth}]{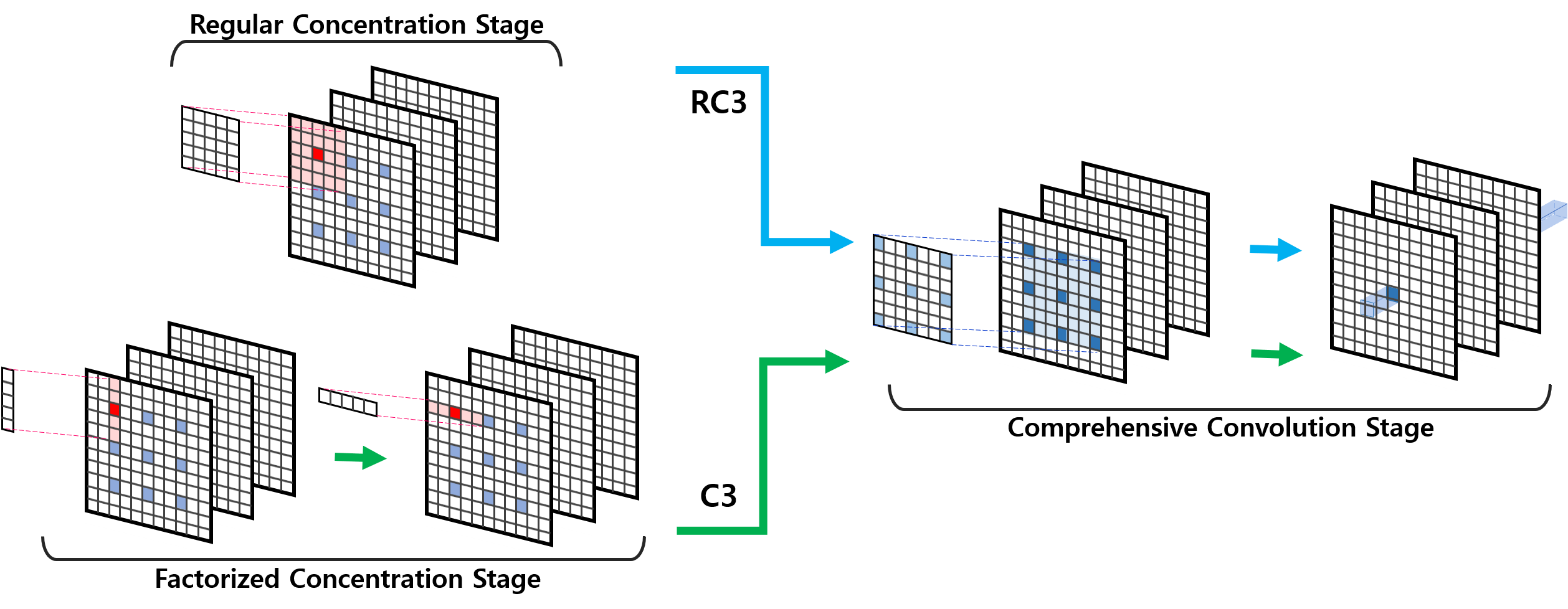}\\
 (a) Structure of our proposed method
     \caption{Structure of Concentrated-Comprehensive Convolutions  block with regular depth-wise convolution (RC3) and Structure of  Concentrated-Comprehensive Convolutions by factorizing concentration stage (C3).}
  \label{fig:CCCblock}%
\end{figure*}

A direct combination of dilated and depth-wise separable convolution skips neighboring information compared to standard convolutional operation and sometimes leads to degradation of accuracy.
 First, we analyze the reason for the disadvantage of depth-wise separable dilated convolution \textit{(ds-Dilate)} in Section \ref{sec:Prob-ds-d}.
 To resolve this problem, we propose a concentrated-comprehensive convolutions (C3) block that maintains the segmentation performance without immense complexity as shown in Figure \ref{fig:CCCblock}.
We also introduce a C3 module for better lightweight segmentation model by applying C3 block in Section \ref{sec:arch}.

\subsection{Issues on Depth-wise Separable Dilated Convolution}
\label{sec:Prob-ds-d}

A dilated convolution is an effective variant of the traditional convolution operator for semantic segmentation in that it creates a large receptive field without decreasing the resolution of the feature map. 
For further reducing computation, applying a depth-wise separable convolution \textit{(ds-Conv)} to a dilated convolution is the key idea. 
Let $C_{i}, C_{o}, H_{i(o)}, W_{i(o)}, M$ and $N$ respectively denote the number of input \& output channels, the height \& width of the input (output) feature map, and the height \& width of the kernel, then a dilated convolutional layer operates on an input feature map $F\in\mathcal{R}^{C_{i}\times H_{i} \times W_{i}}$ using a convolutional kernel $K\in\mathcal{R}^{C_{i}\times C_{o}\times M\times N}$ with a dilation rate $d$. 
Using the kernel $K$, we compute the output feature map $O\in\mathcal{R}^{C_o\times H_{o} \times W_{o}}$ with a dilation rate $d$ as in Equation \ref{eq:st_d}.
\begin{equation}
O_{c',h,w} = \sum_c\sum_m\sum_n F_{c, h+dm, w+dn}K_{c', c, m, n} .
\label{eq:st_d}
\end{equation}
Then, we can apply the depth-wise convolution to the dilated convolution as in Equation \ref{eq:ds_d}.

\begin{equation}
\begin{aligned}
& F'_{c, h, w} =\sum_m\sum_n F_{c, h+dm, w+dn} K^d_{c, m, n}  \\
& O_{c', h, w} = \sum_c F'_{c, h, w} K^p_{c',c}
\end{aligned}
\label{eq:ds_d}
\end{equation}


Note that the process does not calculate cross-channel multiplication, but takes only the spatial multiplication.
Hence, the kernel $K^d$ is for the depth-wise dilated convolution in the spatial dimension, and $K^p$ denotes a kernel for a $1\times1$ point-wise convolution.
Then, the parameter size is reduced from $k^2 C_i C_o$ to $C_i(k^2 + C_o)$  when the kernel size is $k$, i.e., $M=N=k$. 
The number of floating point operations (FLOPs) is also largely reduced from $2k^2C_i C_o$ to $2C_i(k^2 + C_o)$.



However, we observe that this approximation leads to significant performance degradation compared to the case of using the original dilated convolution.
We also note that the dilated convolution has inherent risks as shown in Figure \ref{fig:dilation}.
First, local information can be missing. 
Since the dilated convolutional operation is spatially discrete according to the dilation rate, the information loss is inevitable.
Second, a dilated convolution skips the neighboring feature information, and thus a remote area across large distances affects the result of segmentation.
This includes an area far away from the target area in calculating the convolution, which leads to wrong segmentation in the small and narrow part. 
For example, the target area in the center of the yellow box is misclassified due to the green boxes beneath as shown in Figure \ref{fig:dilation} (b).
Also, unlike standard convolution, a ds-Conv is independent across channels \cite{howard2017mobilenets, chollet2017xception}, and hence spatial information of other channels does not directly flow \cite{howard2017mobilenets}.
This means that ds-Conv uses the feature map sparse than standard convolution.
From the observation in Figure \ref{fig:teaser} (d), we conjecture that the loss of cross-channel information triggers the mentioned risks of dilated convolution, and degrades the performance.

\subsection{Concentrated-Comprehensive Convolution}
\label{sec:CCC}

We propose a concentrated-comprehensive convolution (C3) to prevent segmentation performance degradation based on the observations mentioned in the previous section.
C3 block consists of a concentration stage and a comprehensive convolution stage. 
The concentration stage aggregates local feature information by using a simple convolutional kernel. 
The comprehensive convolution stage uses a dilation rate to get a large receptive field for global consistency. The point-wise convolution is followed for mixing the channel information.

As in Figure \ref{fig:dilation} (a), feature information of white pixels in the region of $(2d-1) \times (2d-1)$ centered on the red pixel will be lost when executing dilated convolution with a dilation rate $d$. Also, the approximated operation led to the loss of information as mention in Section \ref{sec:Prob-ds-d}.
The concentration stage alleviates this loss of feature information by executing simple depth-wise convolution before dilated convolution. This compresses the skipped feature information and improves local consistency as shown in Figure \ref{fig:CCCblock}. 
We note that as the dilation rate increases, the depth-wise convolution in the concentration stage becomes extremely inefficient. 
In most cases, the dilation rate is up to 16, so it becomes intractable in an embedded system.
Specifically, when we use regular depth-wise convolutions to a feature map with $C_{i}$ channels, the number of parameters is $(2d-1)^2C_{i}$. 
When the size of the output of the feature map is $H_o \times W_o$, the number of FLOPs \footnote {We counted multiplication and addition separately.} becomes  $2(2d-1)^2H_o W_o C_{i}$.  
However, if the convolution kernel $K$ is separable, the kernel can be decomposed as $K = K_{row}*K_{col}$. 
For an $N\times N$ kernel, this separable convolution reduces the computational complexity per pixel from $O(N^2)$ to $2O(N)$
We solve the complexity problem by using two depth-wise asymmetric convolutions instead of a regular depth-wise convolution as shown in Figure \ref{fig:CCCblock}.
Also, we insert a non-linearity (PReLU and Batch normalization) between the asymmetric filters.

Comprehensive convolution stage uses depth-wise dilated convolution to widen a receptive field for global consistency. 
After that, we execute the cross-channel operation with a $1 \times 1$ point-wise convolution. 
Since the second stage of C3 is a kind of ds-Conv as shown in Figure \ref{fig:CCCblock}, a small amount number of additional parameters and calculation are required. 
At the same time, C3 block also makes a wide receptive field for segmentation from the comprehensive convolution stage. 
In summary, the C3 block combines both advantages of the depth-wise separable convolution and the dilated convolution by integrating local and global information properly.
Therefore, although the segmentation network based on C3 block has fewer parameters and less computational complexity than the original network, the proposed block can achieve excellent segmentation performance.

\subsection{C3 module}
\label{sec:arch}

\begin{figure}[t]
    \centering
            \includegraphics[width=0.95\linewidth]{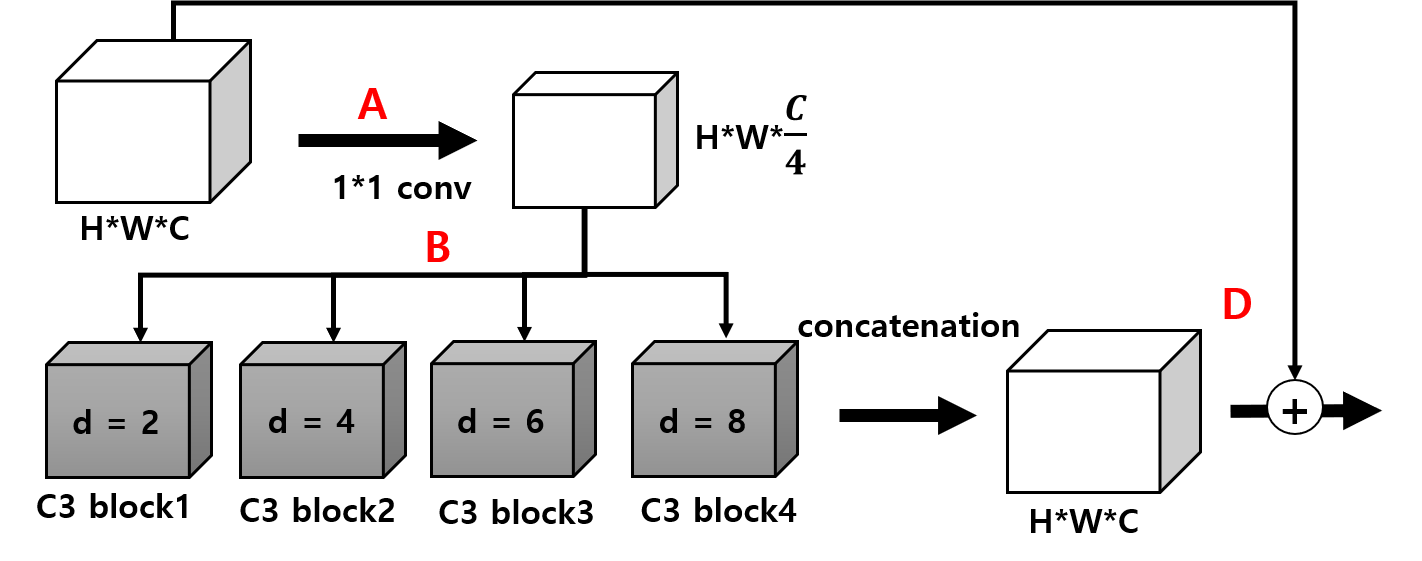}  \\
            (a) C3 module\\
             \includegraphics[width=0.95\linewidth]{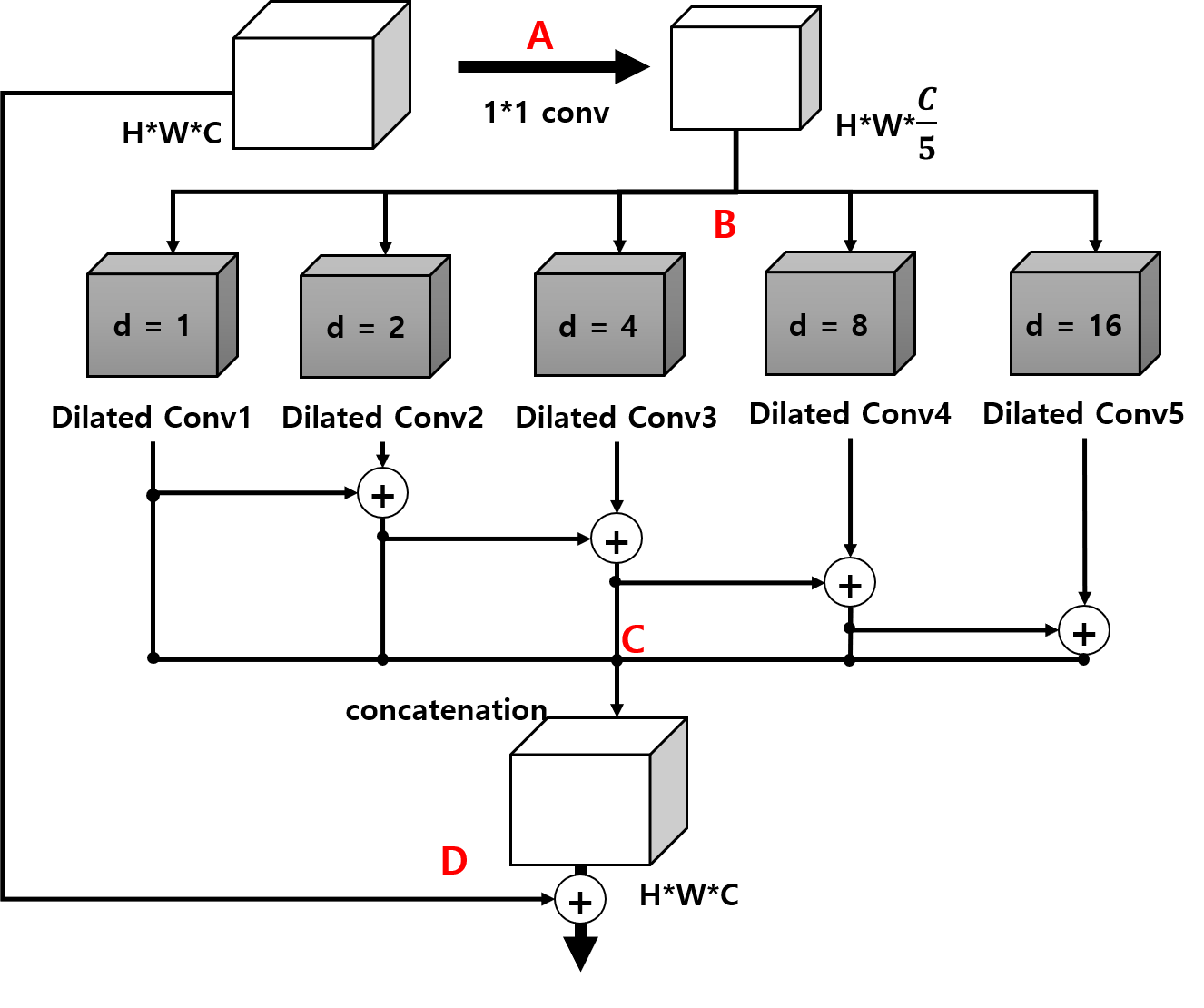}  \\
             (b) ESP module \\
       \caption{\yj{Network} structure of C3 and ESP module.\hj{ A is reducing channel of feature. B is a parallel structure of dilated convolutions. C is hierarchical feature fusion (HFF). D is skip-connection.} }
    \label{fig:CCCmodule}
\end{figure}

\begin{table}[t]
  \centering
     \begin{tabular}{c | c c |  c c}
          & \multicolumn{2}{c|}{ESP module}  & \multicolumn{2}{c}{C3 module} \\
          & Param &FLOPs(M) & Param  & FLOPs(M) \\
          \hline
           \hline
    A & 3,200 & 104.9 & 4,096  & 134.2 \\
     \hline
    B & 28,800 & 943.7 & 9,472  & 299.9 \\
     \hline
    C & - & 0.066& -    & - \\
     \hline
    D & - & 0.016& -    & 0.016 \\
    \hline  \hline
    Total & 31,325      &   1,048.8   & 13,568  & 434.13 \\
    \end{tabular}%
     \caption{Comparison of the number of parameters and FLOPs between ESP and C3 module, under condition of $128 \times 128 \times 128 $ input and output feature map. \nj{A-D are the operations in Figure \ref{fig:CCCmodule}.}}
  \label{tab:modue_compare}%
\end{table}%
We describe C3 module by applying C3 blocks for better lightweight segmentation model.
The original ESP module, which has a parallel structure of $n$ dilated convolutions, each module has dilation rates $2^{i-1}, i=1, \cdots, n$. 
First, it reduces the number of channels in the input feature map by $1/n$ times, and then uses the parallel structure of the extended convolution to the reduced input feature map.
The outputs of each dilated convolutions are element-wise summed hierarchically for degridding\yj{, and this process is referred to} HFF as shown in Figure~\ref{fig:CCCmodule}(b). 

Likewise, when the number of C3 blocks is $n'$, we first reduce the number of channels by $1/n'$ times and then, we apply $n'$ parallel C3 blocks to the reduced feature map as shown in Figure~\ref{fig:CCCmodule}(a).
Unlike other studies \cite{yang2018DenseAspp, mehta2018espnet, Yu2017Drn}, we just concatenate the feature maps without additional post-processing.
Furthermore, ESPNet set $n=5$ including the operation of dilation rate of $1$, but we excluded it to reduce computation. 
This exclusion is reasonable because the concentration stage already gathers neighboring information in C3 block.
Therefore, we set $n'=4$ to view multiple levels of spatial information. 
When the largest dilation rate is $d$, the receptive field of the module is $(2d+1)^2$ in a $3\times 3$ kernel as in the case of ESP module. 
Finally, under the condition of $ 128 \times 128 \times 128 $ input and output feature map, the number of parametesr reduced from $31,325$ to $13,568$, and the number of FLOPs(M) reduced from $1,048.8$ to $434.13$. The detailed calculation is explained in Table~\ref{tab:modue_compare}. 

We note that C3 block can be adapted not only to ESPNet but also to other network based on dilated convolution for making a more efficient model.
By changing the dilated convolutions in the segmentation models to the proposed C3 block, we can reduce the number of parameters and computational cost as well as enhancing the segmentation performance compared to the original network. 
Detailed results will be provided in Section \ref{sec:Exp}.



\begin{table*}[ht]
\centering
    \begin{tabular}{c|l|ccccc}
     & & HFF & Dilate Rate &  Param(M) &  FLOPs(G) & mIOU \\
          \hline  \hline
    1&   Baseline ESPNet & O & 1, 2, 4, 8, 16& 0.364 & 9.07 & 60.74 \\
    2&  Depth-wise separable ESPNet &  O &   1, 2, 4, 8, 16 & 0.128 & 4.93 & 57.81 \\
    \hline
    3&  Depth-wise separable in C3 module  & X & 2, 4, 8, 16 & 0.152 & 5.44 & 54.24 \\

    4&  Depth-wise separable +L in C3 module &  X & 2, 4, 8, 16 & 0.187 & 6.26 & 55.22 \\
    5&  Depth-wise separable +C in C3 module &  X & 2, 4, 8, 16 & 0.181 & 6.41 & 55.52 \\
    6&   with regular Concentration stage in C3 module (RC3) & X & 2, 4, 8, 16 & 0.580 & 15.68 & 58.56 \\
    \hline
    
    7 & with factorized Concentration stage in C3 module (C3) & X & 2, 4, 8, 16 & 0.198 & 6.45 & 60.98 \\
   
    \end{tabular}%
     \caption{C3: Concentrated-Comprehensive Convolution. +L(C): increase the number of layer(channel) in encoder structure for similar parameters and FLOPs. All results are reproduced in the PyTorch framework under the same data augmentation and setting. The input size is $1024\times 512$ for calculating FLOPs. All mIOU results are from Cityscapes benchmark with the test set. }
  \label{tab:ablation}%
\end{table*}%

\begin{figure*}[t]
  \centering
    \begin{tabular}{ccccc}
    \hspace{-2mm}
    \includegraphics[width=0.19\linewidth]{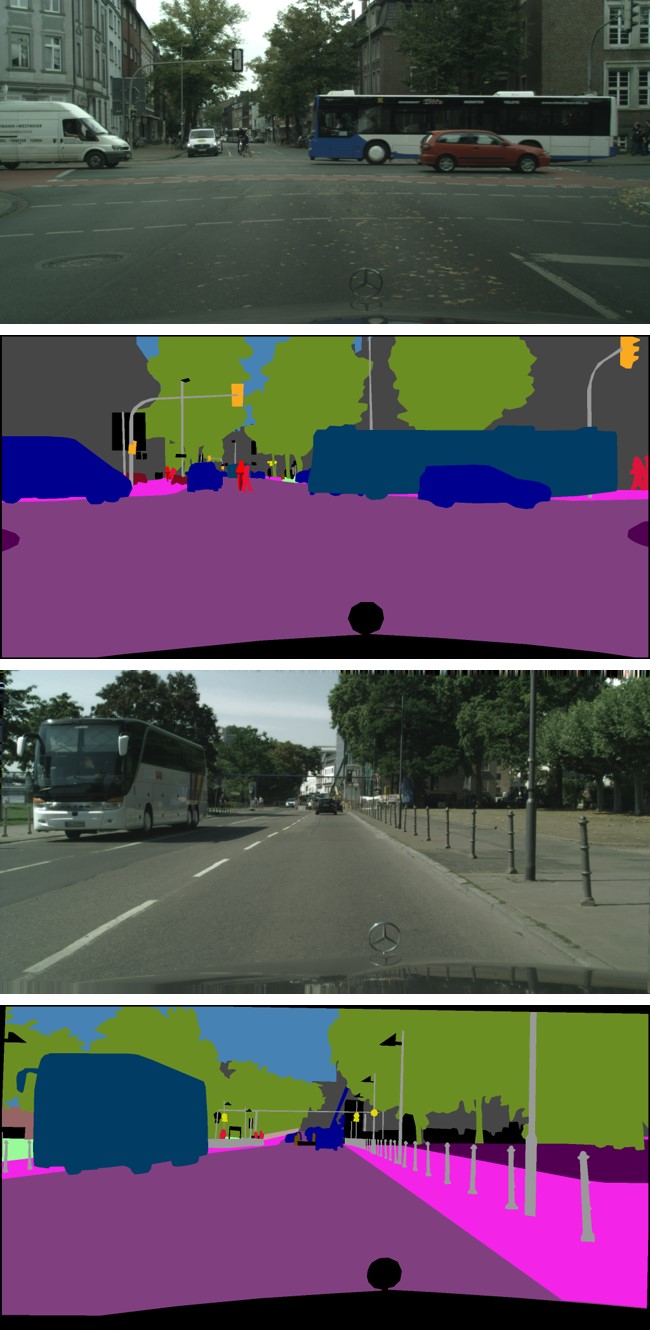}    &
    \hspace{-4mm}
    \includegraphics[width=0.19\linewidth]{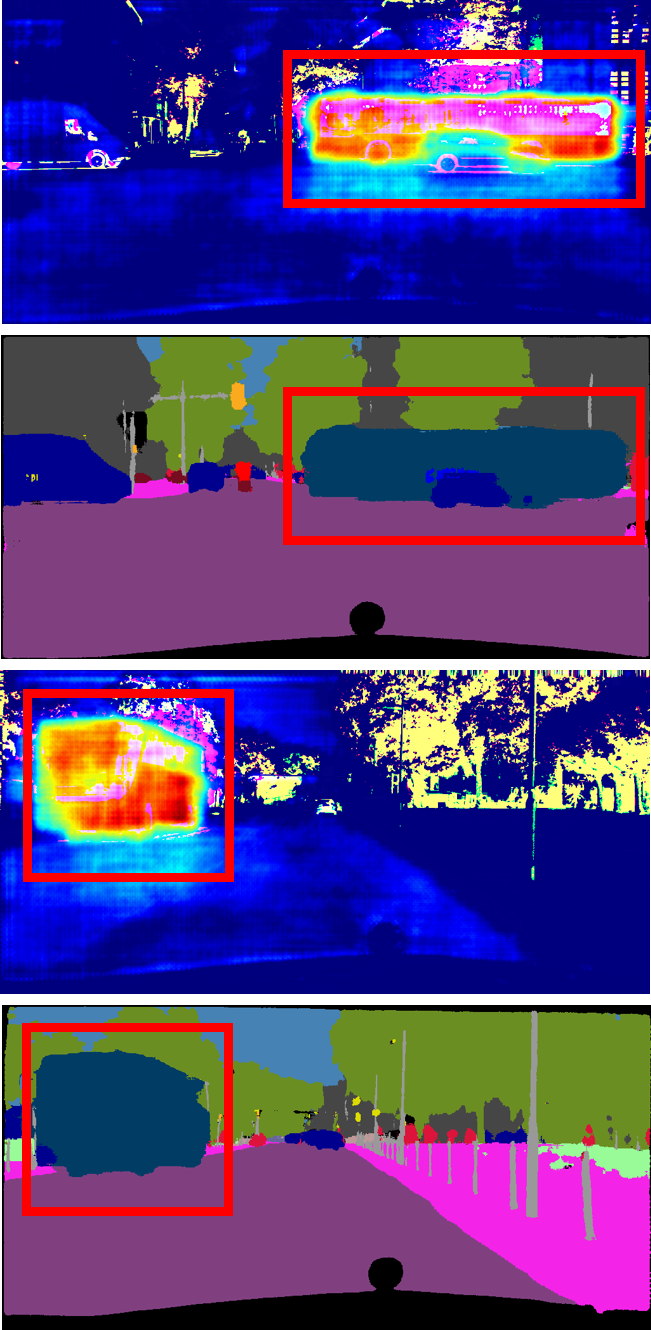}  &
    \hspace{-4mm}
    \includegraphics[width=0.19\linewidth]{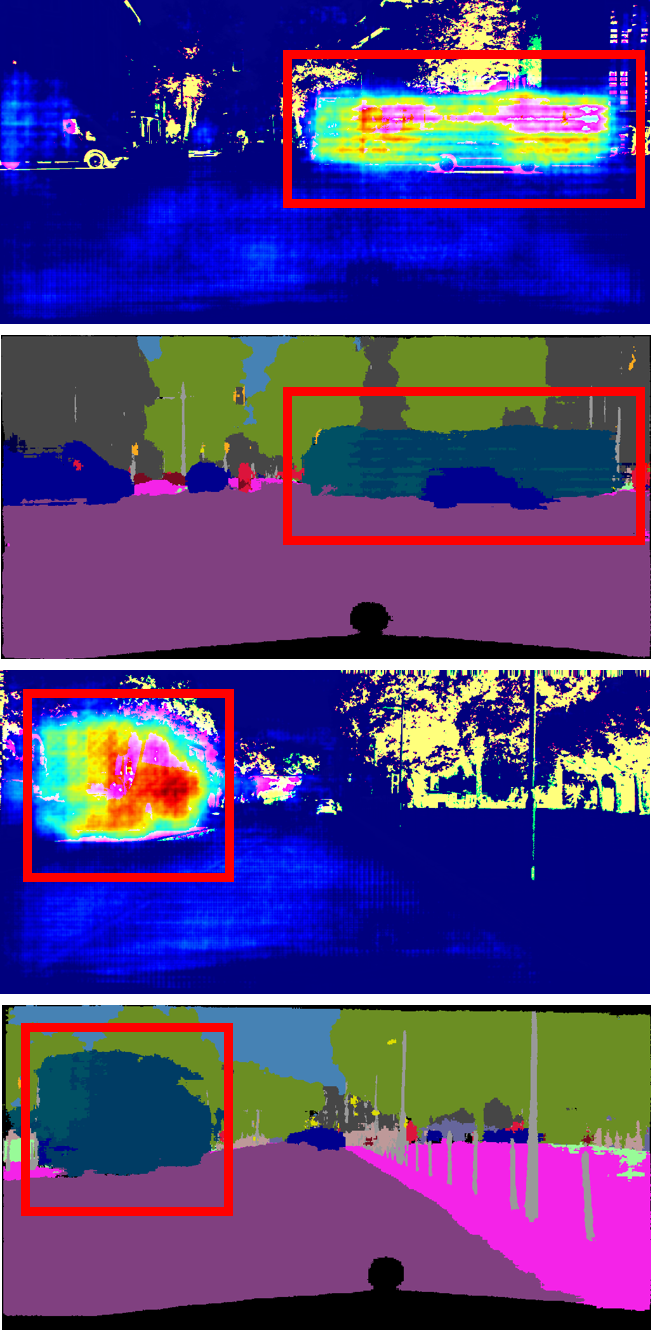}  & 
    \hspace{-4mm}
    \includegraphics[width=0.19\linewidth]{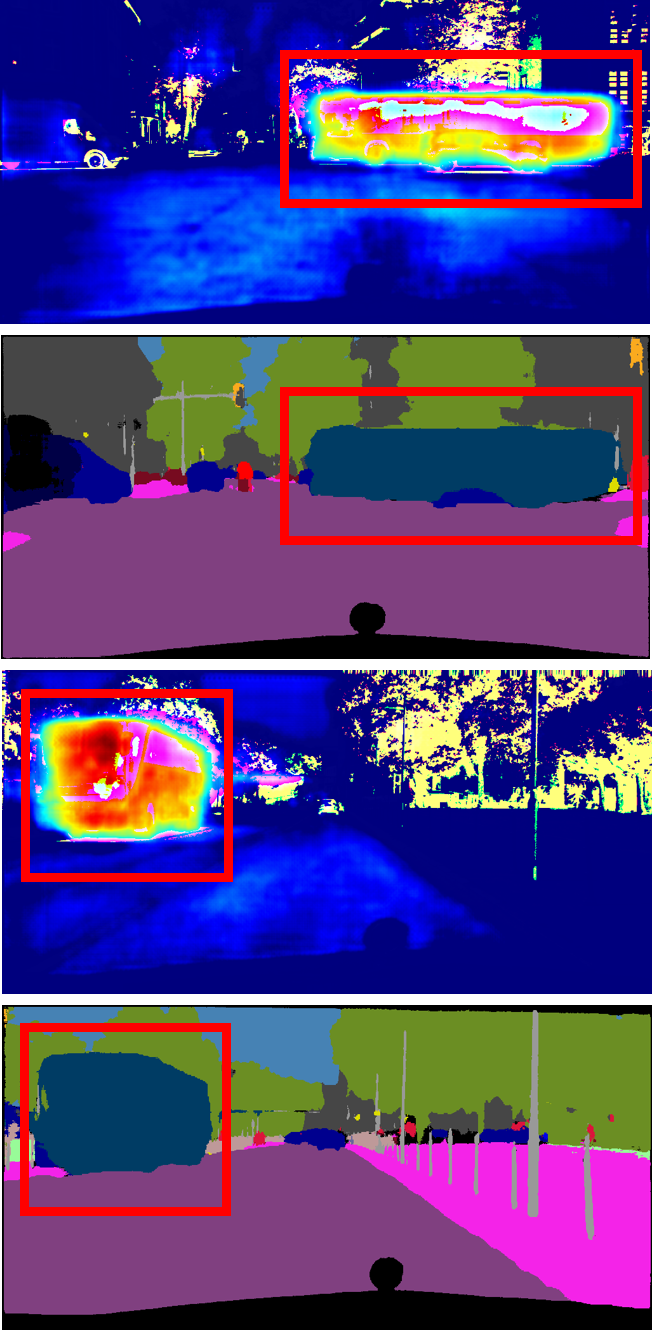}  & 
    \hspace{-4mm}
    \includegraphics[width=0.19\linewidth]{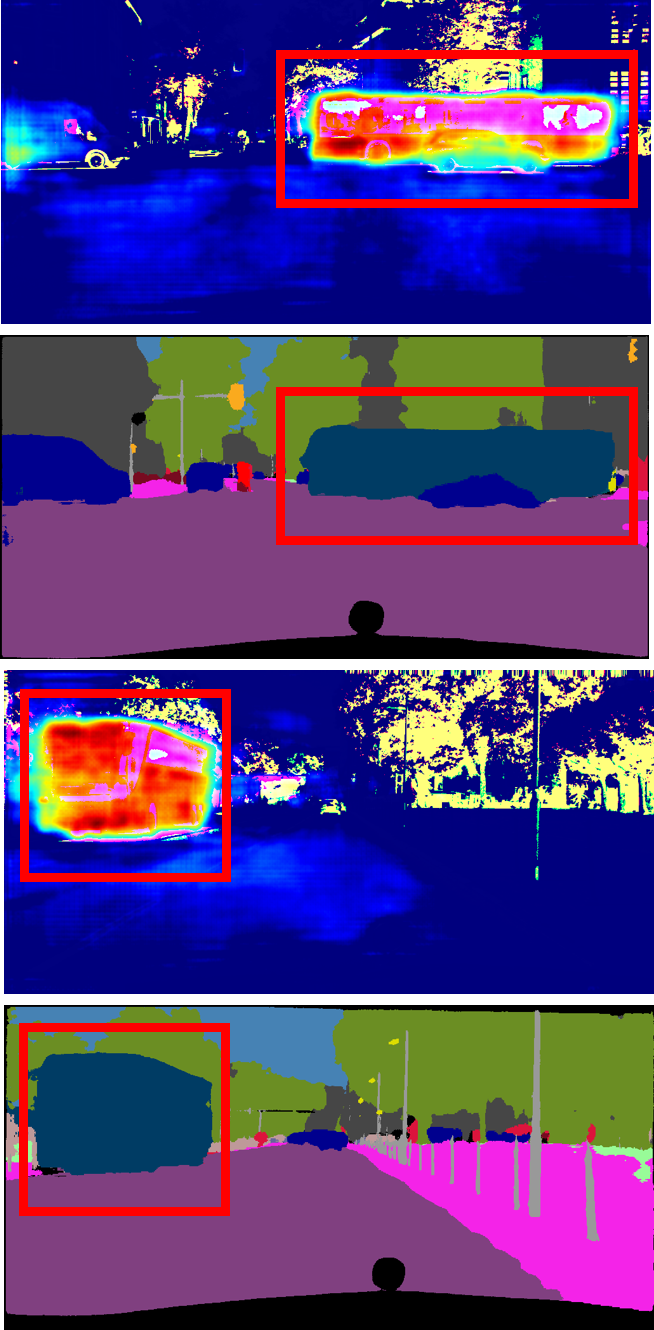} 
   \\
  (a) Input Images & \hspace{-4mm} (b) Experiment 1 & \hspace{-4mm} (c) Experiment 4 & \hspace{-4mm} (d) Experiment 6 & \hspace{-4mm} (e) Experiment 7 \\
    \end{tabular}%
    \caption{The visualization of heatmap from ablation study results. Table \ref{tab:ablation} experiments.
    (a) Experiment 1 is the result of original ESPNet. (b) Experiment 4 is the result of C3 module without a concentration stage (c) Experiment 6 is the result of C3 module with a regular concentration stage. (d) Experiment 7 is the result of C3 module with a factorized concentration stage. From (c) to (d) and (e), the less activated part is enhanced like (b).}
  \label{fig:act_map}%
\end{figure*}%

%% file: Eng/Experiment.tex
\begin{table*}[t]
  \centering
    \begin{tabular*}{0.99\textwidth}{@{\extracolsep{\fill}} l|cccc}
     Method     & Extra dataset & Param(M) & FLOPs(G)  & Class mIOU \\
    \hline
    \hline
    
   
    DRN-A50  \cite{Yu2017Drn}& ImageNet &  23.55 & 400.15 & 67.3 \\
    DRN-C26  \cite{Yu2017Drn}& ImageNet & 20.62 & 355.18  & 68.0 \\
   
    {\color{blue}C3-DRN-A50 }&    {\color{blue}ImageNet}  &   {\color{blue}16.79} &   {\color{blue}289.28} &  {\color{blue} 67.1} \\
    {\color{blue}C3-DRN-C26} & {\color{blue}ImageNet} &  {\color{blue}7.34}& {\color{blue}137.44} & {\color{blue}67.6} \\
    
    \hline
    ENet \cite{paszke2016enet}& no & 0.364 & 8.52 & 58.3 \\
    {\color{blue}C3-ENet} &  {\color{blue}no} &  {\color{blue}0.303} & {\color{blue}6.30} & {\color{blue}60.4} \\
    \hline
    ERFNet  \cite{romera2018erfnet}& no  & 2.10 & 53.48 & 68.0 \\
    {\color{blue}C3-ERFNet} & {\color{blue}no}  & {\color{blue}1.45} & {\color{blue}43.34} &{\color{blue} 69.0} \\
    \hline
    ESPNet \cite{mehta2018espnet} & no &  0.364 & 9.67 & 60.3\\
    ESPNet-tiny  \cite{mehta2018espnet} & no & 0.202 & 6.48 & 56.3\\
   {\color{red} C3Net1 ($d={2, 4, 8, 16}$)} & {\color{red}no} & {\color{red}0.198} & {\color{red}6.45} & {\color{red}61.0} \\
   {\color{red} C3Net2 ($d={2, 3, 7, 13}$) }& {\color{red}no}&{\color{red}0.192} & {\color{red}6.29} & {\color{red}62.0} \\
   {\color{red} C3-ESPNet ($d={1, 2, 4, 8, 16}$ with HFF )}&  {\color{red}no}& {\color{red}0.21} & {\color{red}6.75} & {\color{red}61.1} \\

    \end{tabular*}%
  
  \caption{ We referred the performance of the existing approaches reported in their original papers on Cityscapes benchmark. The FLOPs was calculated on $1024 \times 512$ resolution.}
  \label{tab:result_seg}%
\end{table*}
  

\begin{figure*}[t]
  \centering
    \begin{tabular}{cccc}
    \hspace{-2mm}
    \includegraphics[width=0.23\linewidth]{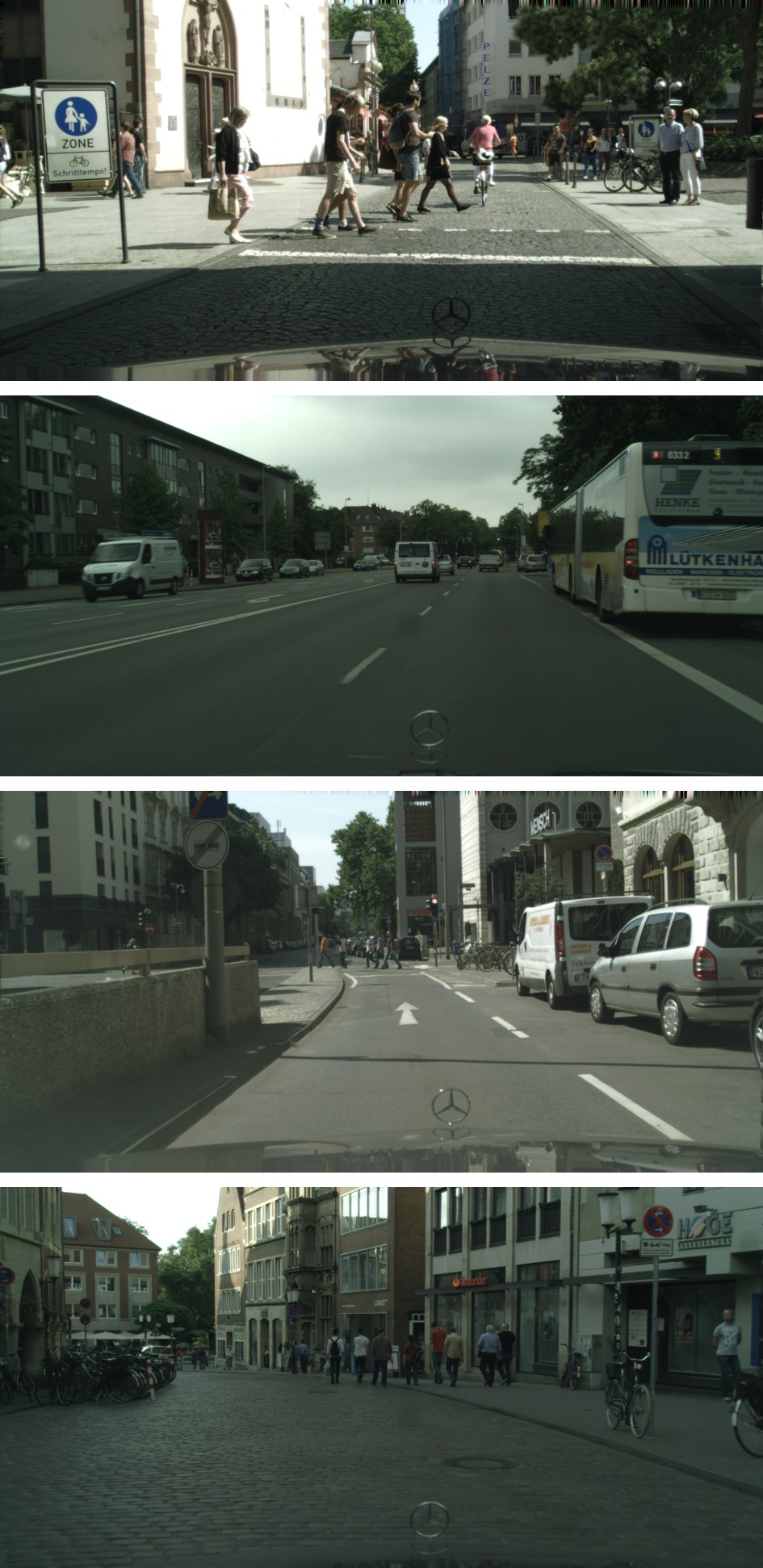}    &
    \hspace{-4mm}
    \includegraphics[width=0.23\linewidth]{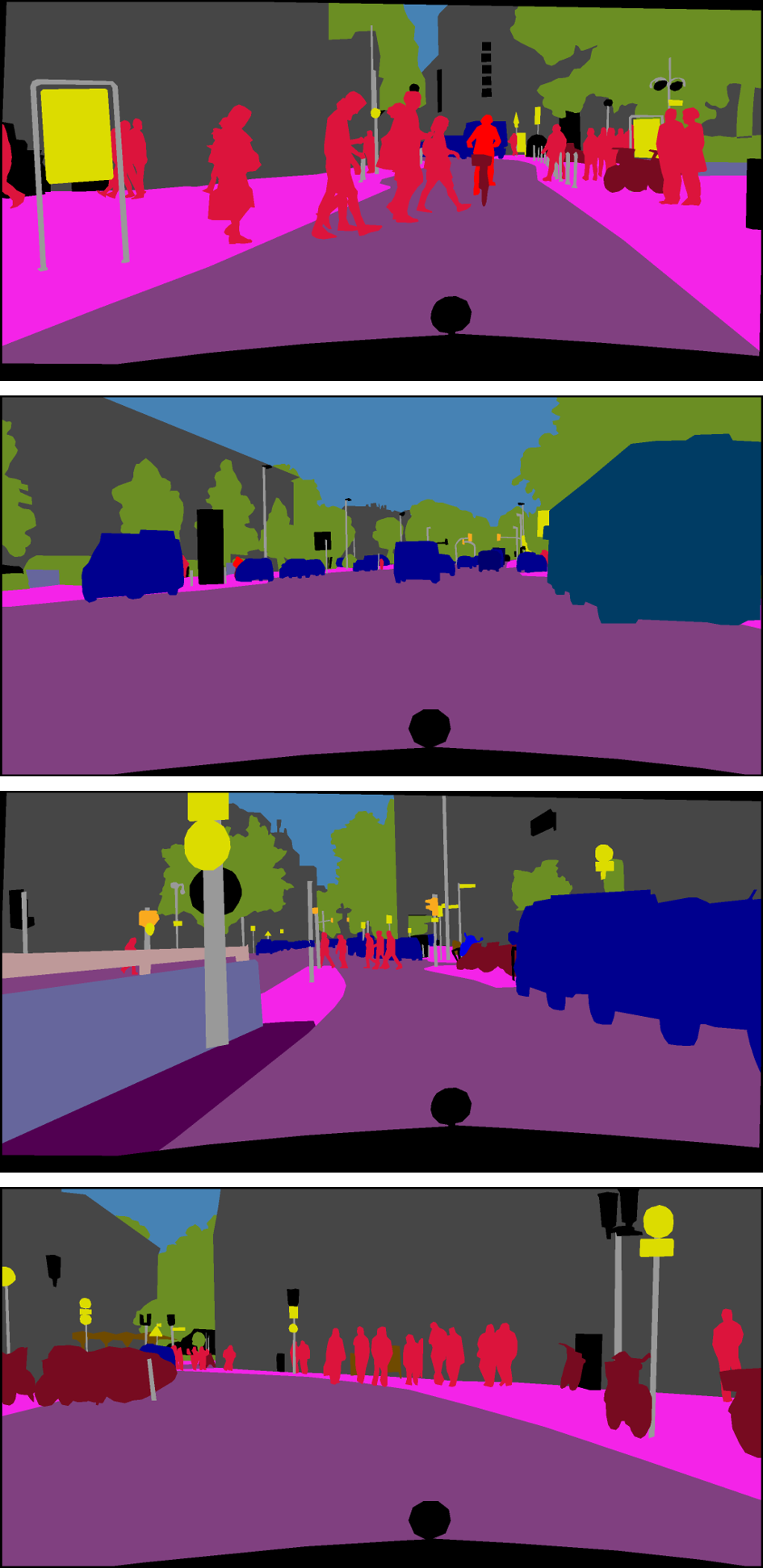}  &
    \hspace{-4mm}
    \includegraphics[width=0.23\linewidth]{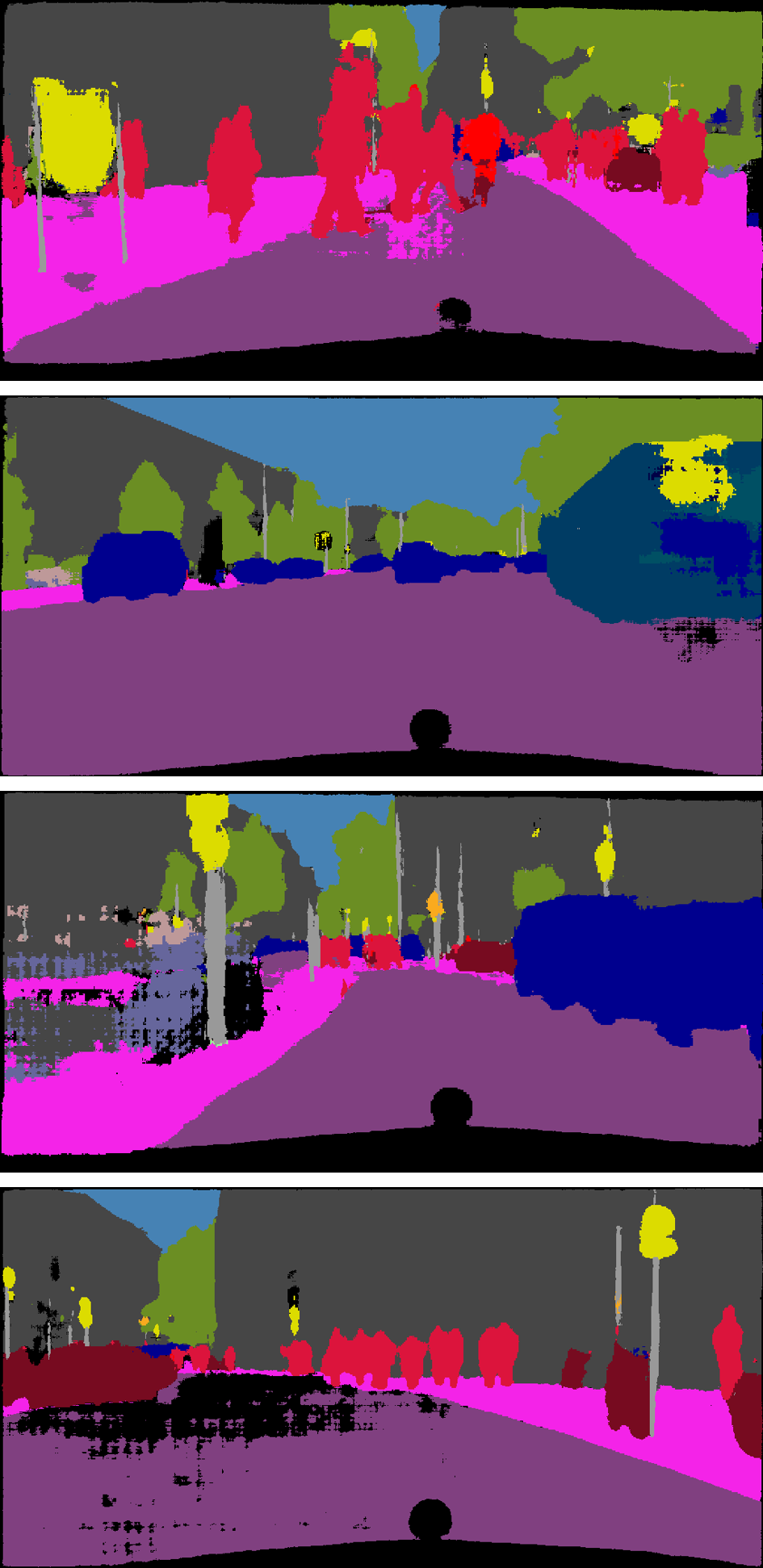}  & 
    \hspace{-4mm}
    \includegraphics[width=0.23\linewidth]{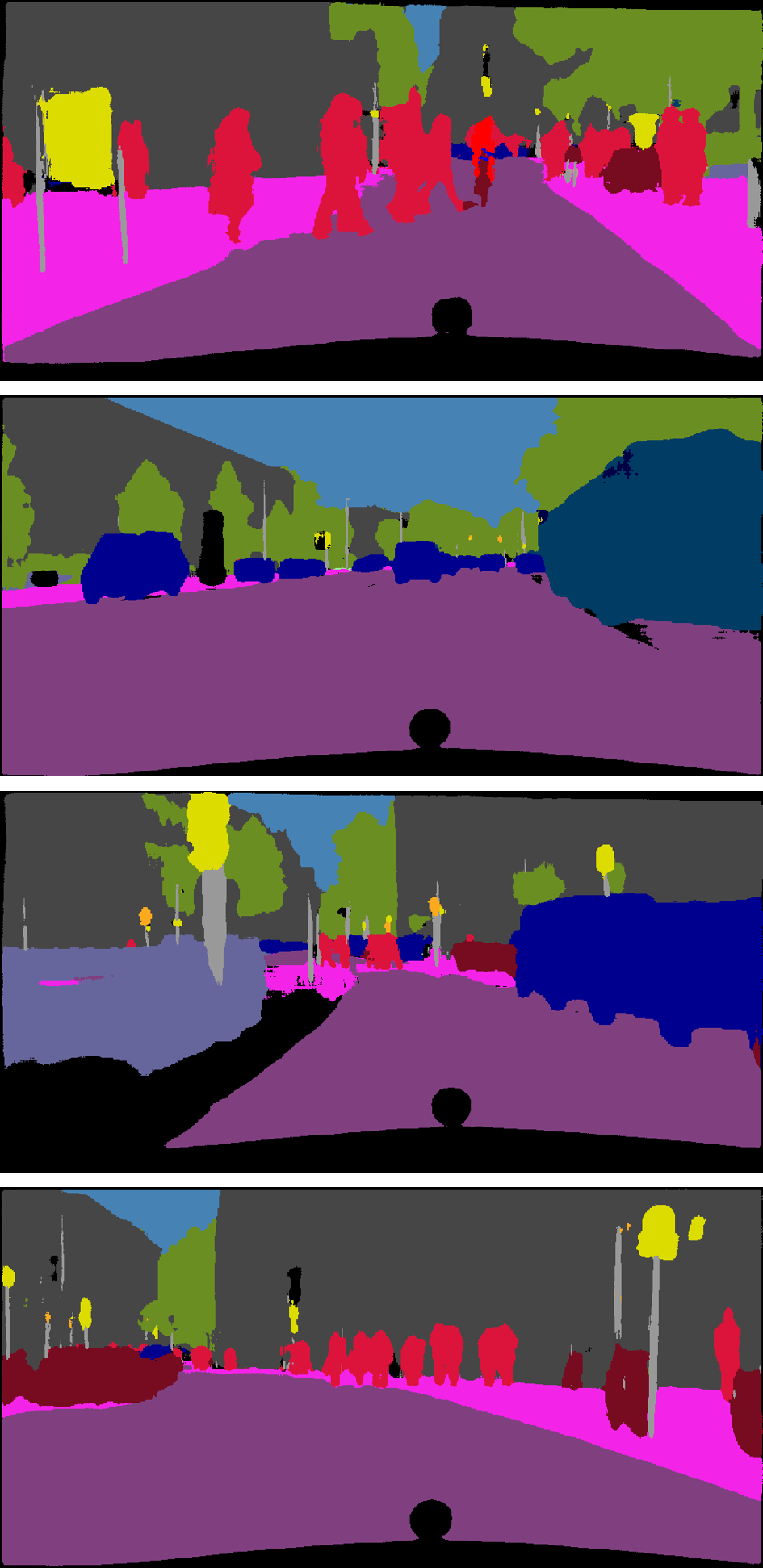}  \\
  (a) Input Images & \hspace{-4mm} (b) Ground-truths & \hspace{-4mm} (c) DS-ESPNet & \hspace{-4mm} (d) C3Net1(ours)\\
    \end{tabular}%
    \caption{Qualitative comparison results on Cityscape validation dataset.}
  \label{fig:result_city}%
  \vspace{-2mm}
\end{figure*}%
  \vspace{-2mm}

\section{Experiment}
\label{sec:Exp}

We evaluated the proposed method and performed an ablation study on Cityscapes, which are widely used in semantic segmentation. 
All the performances were measured using  mean intersection over union (mIOU), the number of parameters, and FLOPs. \footnote {We counted multiplication and addition separately, and we also measured other operations such as activation function, batch normalization, deconvolution, bilinear upsampling, and all that sort of things.}
Also, we showed a classification task on ImageNet\cite{russakovsky2015imagenet} using the C3 block.

\subsection{Ablation Study on C3 block}
\label{exp:ablation}
\noindent 
\textbf{Experimental Setting: }
We show ablation results of the proposed C3 block based on C3 module as introduced in Section \ref{sec:Model} on Cityscapes dataset\cite{cordts2016cityscapes}.
This consists of multiple classes based on urban street scenes. The number of train and validation images are 2,975 and 500 respectively. To train the model with the dataset, we followed the standard data augmentation strategy including random scaling, cropping, and flipping.
The learning rate was set to $1\text{e}^{-3}$ and multiplied by $0.5$ at every 150 and 250 epochs (total 300 epochs).
Adam optimizer \cite{kingma2014adam} with a momentum of $0.9$ and the weight decay $2\text{e}^{-4}$ was used for the training.
As in ESPNet, our modified network also adopts the input reinforcement method which concatenates the feature map with the input image at reduced resolution for designing our encoder.
\yj{For the decoder, we replaced the existing ESP module with C3 module.}

\noindent 
\textbf{Experimental Result: }
Experiment (1) is about original ESPNet with dilation rate $d = \{1,2,4,8,16\}$ in each ESP module. 
Experiment (2) uses the same network structure with the original ESPNet, but the standard dilated convolutions are substituted to depth-wise separable dilated convolutions \textit{(ds-Dilate)}.
Experiments (3)-(7) are ablation studies of the proposed method from C3 module. 
Experiment (3) used simple ds-Dilate in the C3 block.
It means that the concentration stage is not existed in C3 module of Experiment (3).
Experiment (4) is increased the number of layers in the encoder structure from (3), and the Experiment (5) is increased the number of channels from the same encoder structure of (3).
Both (4) and (5) is designed for a fair comparison with the proposed method (Experiment (7)) under a similar parameter and FLOPs size.
The Experiment(6) and (7) are designed for comparing between regular and factorized concentration stage with C3 module.
The concentration stage in (6) consists of regular depth-wise convolution \textit(dw-Conv) as shown in Figure~\ref{fig:CCCblock} RC3 branch.
(7) is the proposed C3 block in Figure \ref{fig:CCCblock} C3 branch for reducing computation by factorizing the regular concentration stage into two depth-wise asymmetric convolutions. 

As shown in Table \ref{tab:ablation} (2)-(5), a naive usage of the depth-wise separable architecture brought significant degradation of the performance (about $3 \sim 5 \%$), and even HFF module could not fully resolve the performance degradation in Experiment (2). 
From the experiments (3)-(5), we can conclude that the concentration stage is critical for resolving the accuracy drop from ds-Dilate, regardless of the number of parameters compared with (6) and (7).
However, the number of parameter is quite large in (6).
In Experiment (7), we showed that the proposed model drastically reduce the number of parameters and FLOPs and could achieve better performance than that of Experiment (6), as well.
From Experiment (7), we found that adding non-linearity between the asymmetric convolution can enhance the performance.

Figure \ref{fig:act_map} is a visualized heatmaps of a bus class by Grad-CAM \cite{selvaraju2017grad}.
The visualized heatmaps of model based on ds-Dilate are not activated enough, and the activated pattern looks like a checkerboard (gridding effect).  \yj{Conversely, by proceeding concentration stage, we observed that our method reduced the checkerboard artifacts (see (c)-(e)).} The  result (d) and (e) of the proposed method shows a similar result with ESPNet which uses standard dilated convolution. Also, the factorized method (e) proves the same capacity with the regular dw-Conv method (d).

\subsection{Evaluation on Cityscapes}
\label{exp:city}

\noindent
\textbf{Experimental Result on Cityscapes Dataset: }
We followed the settings in the respective original papers \cite{romera2018erfnet, Yu2017Drn, paszke2016enet, mehta2018espnet}.
Table \ref{tab:result_seg} shows the evaluation results of the baseline segmentation networks, with switching the dilated convolutional block of the models to the proposed C3 block.
Both of C3Net1 and C3Net2  use ESPNet as a baseline network with varying dilation rate $d$, which is $d = \{2,4,8,16\}$ and $\{2,3,7,13\}$, respectively in C3 module.
C3-ESPNet used all the settings including HFF for degridding and a convolution of dilated ratio $d=1$.
The results show that C3Net2 outperforms C3Net1 about $1\%$ with fewer parameters.
This supports the proposition of the previous work \cite{wang2017understanding} that the dilation rates should be coprime.
ESPNet-tiny model is a smaller version of ESPNet from the original paper, which has a similar number of parameters to the proposed C3Net1 and C3Net2.
The result shows that both C3Net1 and C3Net2 outperform ESPNet-tiny with significant margins, $4.6\%$ for C3Net1 and $5.7\%$ for C3Net2.
On Nvidia-TX2 board with $1024\times512$ resolution, C3Net1 and C3Net2 each marked $15.9$ FPS and $16.4$ FPS. Thus, both can be regarded as real-time methods.

To prove the broader applicability of our C3 block, we conducted further experiments with the other segmentation models : ENet, ERFNet, DRN-A50, DRN-C26, which are based on dilated convolution. 
For every model, our method achieved comparable or higher performance with the reduced number of parameters and FLOPs.
Notably, C3-ENet \yj{achieved} around 2\% performance enhancement with 26\% reduced FLOPs, and C3-ERFNet showed more than 1\% performance enhancement with 30\% reduced parameters.
C3-DRN-A50 reduced the number of parameters and FLOPs by about $31\%$ from the baseline DRN-A50, with slight performance drop ($0.2\%$).
In C3-DRN-C26 case, the number of parameters and FLOPs were reduced by more than $60\%$ from the DRN-C26, but the performance drop was only $0.4\%$.
The overall results show that the proposed C3 block can be incorporated with diverse segmentation models, \yj{as well as} preserving original performance.

\subsection{Evaluation on Classification}

\begin{table}[t]
  \centering
    \begin{tabular}{c|cccc}
     Model     & Param & FLOPs(G)  & Top1 & Top5 \\
     \hline \hline
    DRN-A34 & 21.8 & 17.33 & 75.19 & 91.26 \\     
    DRN-A50 & 25.6 & 19.15 & 77.06 & 93.43 \\
    DRN-C26 & 21.1 & 17.0 & 75.14 & 92.45 \\
    DRN-C42 & 31.2 & 25.1 & 77.06 & 93.43 \\
    
    \hline
    C3-DRN-A50 & 18.8 & 13.85 & 76.51 & 93.02 \\
    C3-DRN-C26 &7.85 & 6.59 &73.43 & 91.33 \\
    C3-DRN-C28* &13.11 & 10.72 &74.39 & 91.99 \\
    C3-DRN-C42 & 9.63 & 8.17 & 74.96 &92.31 \\
    C3-DRN-C44* & 14.90 & 12.3 & 75.64 &92.63 \\
   
    \end{tabular}%
  \caption{ImageNet classification accuracy of DRN and our C3-DRN. \hj{* denotes \nj{increased} residual blocks from the basic model \nj{in the right above}.}}
  \label{tab:classiciation}%
  \vspace{-3mm}
\end{table}

\noindent 
\textbf{Experimental Setting: }
\gs{The evaluation is tested \dy{on} ImageNet-1k~\cite{russakovsky2015imagenet} dataset.
Training was performed with SGD with momentum 0.9 and weight decay $0.0001$. 
Learning rate was initially set to $0.1$ and is reduced by a factor of 10 every 30 epochs.
The training was until 120 epochs with \dy{the} same data augmentation method \nj{following} DRN~\cite{Yu2017Drn}. }

\newcolumntype{L}[1]{>{\raggedright\arraybackslash} m{#1} }

\begin{figure}[t]
  \centering




    
    \begin{tabular}{ccc}
    \includegraphics[width=0.3\linewidth]{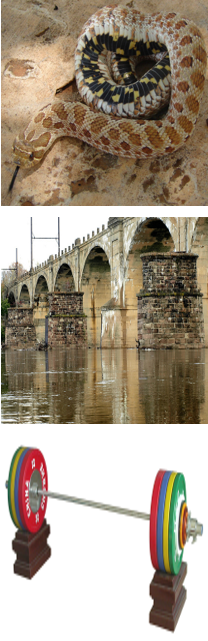}    &
    \includegraphics[width=0.3\linewidth]{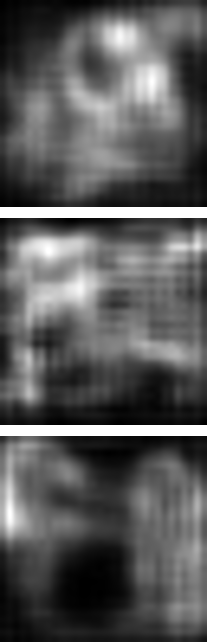}&
    \includegraphics[width=0.3\linewidth]{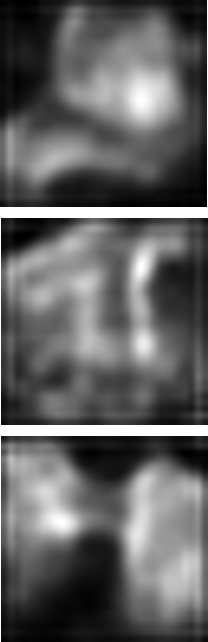}  \\

    (a) \small{Image}   & (b) \small{DRN-C26}   & (c) \small{C3-DRN-C44}  \\
    \end{tabular}%
    
    \caption{Qualitative comparison of visualized activation maps on ImageNet validation dataset}
  \label{fig:DRNact_map}%
\end{figure}%


\noindent
\textbf{ImageNet-1k Benchmark Results: }
\yj{
From the experiments on \nj{the} segmentation task, we have shown that the proposed C3 block can successfully substitute the dilated convolutional block.
One step further from segmentation, we tested the proposed block to (DRN)~\cite{Yu2017Drn} which \nj{incorporate the} dilated convolutional scheme to skip-connection for classification.
\yj{The result in Table~\ref{tab:classiciation} shows that the proposed method can achieve comparable classification accuracy considering that the parameter size was reduced. 
}

For example, the \nj{C3-DRN-C42} reduced the parameters and FLOPss by more than half compared to the DRN-C26, but the top1 accuracy dropped only 0.18\%. Also C3-DRN-C44* reduced the number of parameters by 30\% and the number of FLOPs by half, but better accuracy than DRN-C26. 
\hj{The low resolution feature map degrades the classification accuracy and localization power according to \nj{\cite{Yu2017Drn}}.
}
DRN showed that adding dilated convolution strengthen the localization power and hence can enhance the classification accuracy without adding further parameters.
By changing the dilated convolutional parts in DRN to the proposed C3 block, we tested whether our method can substitute the dilated convolution in this case, as well.}

We visualized the dense pixel-level class activation map from \nj{the DRN and our C3-DRN to show each model's} localization power \nj{in Figure~\ref{fig:DRNact_map}.}
From the figure, we can see that the proposed method generated the activation maps with better resolution than DRN, which means that our method also have enough localization capacity to DRN.

%% file: Eng/Conclusion.tex
\section{Conclusion}
\label{conclusion}
In this work, we proposed Concentrated-Comprehensive Convolutions (C3) block for light weight semantic segmentation. Our C3 block comprises of the information concentration stage and the comprehensive convolution stage both of which are based on two consecutive convolutional blocks. More specifically, the former block improves the local consistency by using two depth-wise asymmetric convolutions by compressing information neighboring pixels, and the latter one increases a receptive field by using a depth-wise separable dilated convolution. Throughout the extensive experiments, it turns out that our proposed method could integrate local and global information effectively and reduce the number of parameters and flops significantly. Furthermore, C3 block is generally applicable to other semantic segmentation models and can be adopted to other tasks such as classification.

%% file: ICCV_C3.bbl
\begin{thebibliography}{10}\itemsep=-1pt

\bibitem{chen2017rethinking}
L.-C. Chen, G.~Papandreou, F.~Schroff, and H.~Adam.
\newblock Rethinking atrous convolution for semantic image segmentation.
\newblock {\em arXiv preprint arXiv:1706.05587}, 2017.

\bibitem{chen2018encoder}
L.-C. Chen, Y.~Zhu, G.~Papandreou, F.~Schroff, and H.~Adam.
\newblock Encoder-decoder with atrous separable convolution for semantic image
  segmentation.
\newblock In {\em ECCV}, 2018.

\bibitem{chollet2017xception}
F.~Chollet.
\newblock Xception: Deep learning with depthwise separable convolutions.
\newblock {\em arXiv preprint}, pages 1610--02357, 2017.

\bibitem{cordts2016cityscapes}
M.~Cordts, M.~Omran, S.~Ramos, T.~Rehfeld, M.~Enzweiler, R.~Benenson,
  U.~Franke, S.~Roth, and B.~Schiele.
\newblock The cityscapes dataset for semantic urban scene understanding.
\newblock In {\em Proceedings of the IEEE conference on computer vision and
  pattern recognition}, pages 3213--3223, 2016.

\bibitem{he2016deep}
K.~He, X.~Zhang, S.~Ren, and J.~Sun.
\newblock Deep residual learning for image recognition.
\newblock In {\em Proceedings of IEEE Conference on Computer Vision and Pattern
  Recognition (CVPR)}, pages 770--778, 2016.

\bibitem{howard2017mobilenets}
A.~G. Howard, M.~Zhu, B.~Chen, D.~Kalenichenko, W.~Wang, T.~Weyand,
  M.~Andreetto, and H.~Adam.
\newblock Mobilenets: Efficient convolutional neural networks for mobile vision
  applications.
\newblock {\em arXiv preprint arXiv:1704.04861}, 2017.

\bibitem{kaiser2018depthwise}
L.~Kaiser, A.~N. Gomez, and F.~Chollet.
\newblock Depthwise separable convolutions for neural machine translation.
\newblock In {\em International Conference on Learning Representations}, 2018.

\bibitem{kingma2014adam}
D.~P. Kingma and J.~Ba.
\newblock Adam: A method for stochastic optimization.
\newblock {\em arXiv preprint arXiv:1412.6980}, 2014.

\bibitem{yang2018DenseAspp}
C.~Z. Z. L. K.~Y. Maoke~Yang, Kun~Yu.
\newblock Denseaspp for semantic segmentation in street scenes.
\newblock In {\em CVPR}, 2018.

\bibitem{paszke2016enet}
A.~Paszke, A.~Chaurasia, S.~Kim, and E.~Culurciello.
\newblock Enet: A deep neural network architecture for real-time semantic
  segmentation.
\newblock {\em arXiv preprint arXiv:1606.02147}, 2016.

\bibitem{poudel2018contextnet}
R.~P. Poudel, U.~Bonde, S.~Liwicki, and C.~Zach.
\newblock Contextnet: Exploring context and detail for semantic segmentation in
  real-time.
\newblock {\em arXiv preprint arXiv:1805.04554}, 2018.

\bibitem{romera2018erfnet}
E.~Romera, J.~M. Alvarez, L.~M. Bergasa, and R.~Arroyo.
\newblock Erfnet: Efficient residual factorized convnet for real-time semantic
  segmentation.
\newblock {\em IEEE Transactions on Intelligent Transportation Systems},
  19(1):263--272, 2018.

\bibitem{russakovsky2015imagenet}
O.~Russakovsky, J.~Deng, H.~Su, J.~Krause, S.~Satheesh, S.~Ma, Z.~Huang,
  A.~Karpathy, A.~Khosla, M.~Bernstein, et~al.
\newblock Imagenet large scale visual recognition challenge.
\newblock {\em International Journal of Computer Vision}, 115(3):211--252,
  2015.

\bibitem{mehta2018espnet}
A.~C. L.~S. Sachin~Mehta, Mohammad~Rastegari and H.~Hajishirzi.
\newblock Espnet: Efficient spatial pyramid of dilated convolutions for
  semantic segmentation.
\newblock In {\em ECCV}, 2018.

\bibitem{sandler2018inverted}
M.~Sandler, A.~Howard, M.~Zhu, A.~Zhmoginov, and L.-C. Chen.
\newblock Inverted residuals and linear bottlenecks: Mobile networks for
  classification, detection and segmentation.
\newblock {\em arXiv preprint arXiv:1801.04381}, 2018.

\bibitem{selvaraju2017grad}
R.~R. Selvaraju, M.~Cogswell, A.~Das, R.~Vedantam, D.~Parikh, and D.~Batra.
\newblock Grad-cam: Visual explanations from deep networks via gradient-based
  localization.
\newblock In {\em Proceedings of the IEEE International Conference on Computer
  Vision}, pages 618--626, 2017.

\bibitem{siam2018rtseg}
M.~Siam, M.~Gamal, M.~Abdel-Razek, S.~Yogamani, and M.~Jagersand.
\newblock Rtseg: Real-time semantic segmentation comparative study.
\newblock In {\em IEEE International Conference on Image Processing (ICIP)},
  pages 1603 -- 1607, 2018.

\bibitem{szegedy2017inception}
C.~Szegedy, S.~Ioffe, V.~Vanhoucke, and A.~A. Alemi.
\newblock Inception-v4, inception-resnet and the impact of residual connections
  on learning.
\newblock In {\em AAAI}, volume~4, page~12, 2017.

\bibitem{szegedy2015going}
C.~Szegedy, W.~Liu, Y.~Jia, P.~Sermanet, S.~Reed, D.~Anguelov, D.~Erhan,
  V.~Vanhoucke, and A.~Rabinovich.
\newblock Going deeper with convolutions.
\newblock In {\em Proceedings of IEEE Conference on Computer Vision and Pattern
  Recognition (CVPR)}, pages 1--9, 2015.

\bibitem{szegedy2016rethinking}
C.~Szegedy, V.~Vanhoucke, S.~Ioffe, J.~Shlens, and Z.~Wojna.
\newblock Rethinking the inception architecture for computer vision.
\newblock In {\em Proceedings of IEEE Conference on Computer Vision and Pattern
  Recognition (CVPR)}, pages 2818--2826, 2016.

\bibitem{vallurupalli2018efficient}
N.~Vallurupalli, S.~Annamaneni, G.~Varma, C.~Jawahar, M.~Mathew, and S.~Nagori.
\newblock Efficient semantic segmentation using gradual grouping.
\newblock In {\em Proceedings of the IEEE Conference on Computer Vision and
  Pattern Recognition Workshops}, pages 598--606, 2018.

\bibitem{wang2017understanding}
P.~Wang, P.~Chen, Y.~Yuan, D.~Liu, Z.~Huang, X.~Hou, and G.~Cottrell.
\newblock Understanding convolution for semantic segmentation.
\newblock In {\em IEEE Winter Conf. on Applications of Computer Vision (WACV)},
  2018.

\bibitem{xie2017aggregated}
S.~Xie, R.~Girshick, P.~Doll{\'a}r, Z.~Tu, and K.~He.
\newblock Aggregated residual transformations for deep neural networks.
\newblock In {\em Computer Vision and Pattern Recognition (CVPR), 2017 IEEE
  Conference on}, pages 5987--5995. IEEE, 2017.

\bibitem{Yu2017Drn}
F.~Yu, V.~Koltun, and T.~Funkhouser.
\newblock Dilated residual networks.
\newblock In {\em Computer Vision and Pattern Recognition (CVPR)}, 2017.

\bibitem{zhang2017shufflenet}
X.~Zhang, X.~Zhou, M.~Lin, and J.~Sun.
\newblock Shufflenet: An extremely efficient convolutional neural network for
  mobile devices.
\newblock {\em arXiv preprint arXiv:1707.01083}, 2017.

\end{thebibliography}
